\documentclass[conference]{IEEEtran}

\usepackage[english]{babel}

\RequirePackage[l2tabu, orthodox]{nag}

\usepackage{pdflscape}

\usepackage{newtxtext,newtxmath}

\usepackage{cite}

\usepackage{booktabs}
\newcommand{\ra}[1]{\renewcommand{\arraystretch}{#1}}

\usepackage{multirow}

\usepackage{caption}
\usepackage{float}
\usepackage{tabularx}
\usepackage{xcolor}
\usepackage{needspace}
\usepackage[pdftex]{graphicx}
\graphicspath{{./figures/}{./resources/}}
\DeclareGraphicsExtensions{.pdf,.png,.jpg,.jpeg}
\usepackage{wrapfig}
\usepackage{tikz}
\usetikzlibrary{shapes,arrows}
\usepackage{pgfplots}
\pgfplotsset{compat=1.9}

\usepackage{amsmath}
\usepackage{mathtools}

\ifCLASSOPTIONcompsoc
  \usepackage[caption=false,font=normalsize,labelfont=sf,textfont=sf]{subfig}
\else
  \usepackage[caption=false,font=footnotesize]{subfig}
\fi

\usepackage{url}
\usepackage{hyperref}

\usepackage{xspace}
\makeatletter
\DeclareRobustCommand\onedot{\futurelet\@let@token\@onedot}
\def\@onedot{\ifx\@let@token.\else.\null\fi\xspace}

\def\etal{\emph{et al}\onedot}
\makeatother

\begin{document}
\title{Towards Sustainable Deep Learning for Wireless Fingerprinting Localization}

\author{\IEEEauthorblockN{%
    An\v{z}e~Pirnat\IEEEauthorrefmark{1},
    Bla\v{z}~Bertalani\v{c}\IEEEauthorrefmark{1},
    Gregor~Cerar\IEEEauthorrefmark{1},
    Mihael~Mohor\v{c}i\v{c}\IEEEauthorrefmark{1},
    Marko~Me\v{z}a\IEEEauthorrefmark{2},
    and Carolina~Fortuna\IEEEauthorrefmark{1}
}\IEEEauthorblockA{%
    \IEEEauthorrefmark{1}Department of Communication Systems, Jo\v{z}ef Stefan Institute, Slovenia.\\
    \IEEEauthorrefmark{2}Faculty of Electrical Engineering, University of Ljubljana, Slovenia.
}%
ap6928@student.uni-lj.si, blaz.bertalanic@ijs.si, gregor.cerar@ijs.si, miha.mohorcic@ijs.si, \\marko.meza@fe.uni-lj.si, carolina.fortuna@ijs.si
}

%



\maketitle

\begin{abstract}
Location based services, already popular with end users, are now inevitably becoming part of new  wireless infrastructures and emerging business processes. The increasingly popular Deep Learning (DL) artificial intelligence methods perform very well in wireless fingerprinting localization based on extensive indoor radio measurement data. However, with the increasing complexity these methods become computationally very intensive and energy hungry, both for their training and subsequent operation. Considering only mobile users, estimated to exceed 7.4 billion by the end of 2025, and assuming that the networks serving these users will need to perform only one localization per user per hour on average, the machine learning models used for the calculation would need to perform $65 \times 10^{12}$ predictions per year. Add to this equation tens of billions of other connected devices and applications that rely heavily on more frequent location updates, and it becomes apparent that localization will contribute significantly to carbon emissions unless more energy-efficient models are developed and used. This motivated our work on a new DL-based architecture for indoor localization that is more energy efficient compared to related state-of-the-art approaches while showing only marginal performance degradation. A detailed performance evaluation shows that the proposed model produces only 58\,\% of the carbon footprint while maintaining 98.7\,\% of the overall performance compared to state of the art model external to our group. Additionally, we elaborate on a methodology to calculate the complexity of the DL model and thus the CO$_{\text{2}}$ footprint during its training and operation.
\end{abstract}


%
\IEEEpeerreviewmaketitle

\section{Introduction}
\label{sec:intro}






Location-based services (LBS) are software services that take into account a geographic location and even context of an entity~\cite{junglas2008location} in order to adjust the content, information or functionality delivered. Entities can be people, animals, plants, assets and any other object. Perhaps the most widely used LBS is the Global Positioning System (GPS), which integrates data from satellite navigation systems and cell towers \cite{hofmann2012global} and is used daily in navigation systems. Another popular application of LBS is locating tagged items and assets in indoor environments. 

With 5G systems, accurate localization is no longer only important for the provision of more relevant information to the end user, but also for optimal operation and management of the network, e.g. for creating and steering the beams of antenna array-based radio heads \cite{kanhere2021position}. As discussed in \cite{kanhere2021position}, the poor performance of fundamental geometry-based techniques in challenging indoor environments characterized by non line-of-sight (NLoS) and/or multipath propagation can be significantly improved by using higher mmWave frequency bands and steerable multiple-input multiple-output (MIMO) antennas along with advanced techniques such as cooperative localization, machine learning (ML) and user tracking. Given the ubiquitous presence of wireless networks and the associated availability of radio-frequency (RF) measurements, ML methods promise the highest accuracy, albeit at a higher deployment cost. In particular, in the offline training phase, ML methods use available RF measurements to create a fingerprint database of the wireless environment, hence we refer to this localization approach as wireless fingerprinting. The fingerprint database is then used in the online localization phase to compare the real-time RF measurement with the stored (measured or estimated) values associated with exact or estimated locations. 

Recent advances in Deep Learning (DL) \cite{yan2021extreme} have enabled particularly accurate localization, and such models trained with large amounts of data are considered the most promising enablers for the future LBS. However, the development and use of DL models involves additional technical complexity, increased energy consumption and corresponding environmental impacts. Recently, the impact of such technologies has received increased attention from regulators and the public, triggering related research activities \cite{strubell2019}. One way to reduce the environmental impact of power-hungry AI technology is to increase the proportion of electricity from clean energy sources such as wind, solar and hydro. However, this must be complemented by further efforts to optimize energy consumption relative to the performance of existing and emerging technologies. Studies on estimating the energy consumption of ML models~\cite{Garcia2019} show that the increasing complexity of models, manifested in the number of weights, the type of layers and their respective parameters, affects both their performance and energy efficiency. In DL architectures, one way to optimize the use of energy is to reduce the size of the filters, also referred to as kernels, that represent matrices used to extract features from the image. In these filters, we can adjusts the amount of movement over the image by a stride. Another way is to adjust pools, which represent layers that resize the output of a filter and thus reduce the number of parameters passed to subsequent layers, making a model lighter and faster.

In this paper, we propose a new DL architecture that is an adaptation of ResNet18 for the indoor localization problem under consideration, and prove that its performance is comparable to the state of the art while being much more energy efficient. Our contributions are as follows:
\begin{itemize}
    \item We design a new model with a kernel and a max pool that extract the most useful information from the subcarriers of a single antenna (e.g. those with the highest response), thus taking a different approach from the square kernel and pools of ResNet18.
    \item We elaborate on a methodology for computing computational complexity during training and operation of a DL architecture. 
    \item In a carbon footprint study we show that the proposed DL model during training produces only 58\,\% of the carbon footprint and maintains 98.7\,\% of the overall performance. 
\end{itemize}

The paper is organized as follows. Section ~\ref{sec:related_work} describes related work, Section~\ref{sec:setup} provides the problem statement, Section~\ref{sec:model} presents the proposed model and elaborates on  methodological aspects, while Section~\ref{sec:evaluation} provides a comprehensive evaluation. Finally, Section~\ref{sec:conclusion}  concludes the paper.

\section{Related Work}
\label{sec:related_work}
In this section, we first present some very recent related work on wireless fingerprinting for massive MIMO setups using deep learning, and then summarize the state of the art in energy-efficient design of ML models.  

\subsection{ML-based wireless fingerprinting}
\label{subsec:fingerprinting}
In~\cite{DeBast2020}, De Bast~\textit{et al.} proved that convolutional neural networks (CNNs) can be effectively used for wireless fingerprinting, and that more antennas significantly increase localization accuracy. They designed a model for a massive MIMO setup with 64 antennas, using stride 1\,$\times$\,n (1D convolution). The model includes 13 convolutional layers and 3 dense layers, enhanced by skip-connection and drop-out layers. In~\cite{Pollin2020}, De Bast~\textit{et al.} proposed another model based on the dense convolutional network (DenseNet)~\cite{Huang2017} and evaluated its performance under LoS and NLoS conditions. They concluded that in addition to the direct signal paths, the model needs to exploit the multipath components for more robust and accurate positioning. The proposed  dense blocks consist of 4 convolutional layers, skip-connection layers and concatenation layers. After each convolutional layer there is an average pooling layer and a batch normalization layer. 

Using the same dataset as~\cite{DeBast2020}, Widmaier~\textit{et al.}~\cite{Widmaier2019} showed that objects that are in the line of sight can be localized better than those that are not. They also proved the robustness of their model by running it for a few days without any noticable decrease in accuracy.

Cerar~\textit{et al.}~\cite{cerar2021improving} came to the same conclusion as~\cite{DeBast2020}, but on a different data set with a smaller MIMO array of 16 antennas. However, since they use 4 times fewer antennas than~\cite{DeBast2020},~\cite{Widmaier2019} and~\cite{Pollin2020}, their results are less accurate. 

Chin~\textit{et al.}~\cite{chin2020intelligent} has proven that it is possible to use a MIMO setup with 16 antennas and channel state information (CSI) to fingerprint in the case of shielding, where GPS would not work. They have also shown how effective convolutional layers are compared to fully connected layers in DL. Also Sobehy~\textit{et al.}~\cite{Sobehy2020} designed their model for a MIMO setup with 16 antennas and used the k-nearest neighbors algorithm for wireless fingerprinting based on CSI, proving that the most reliable input feature is the magnitude. 

Arnold~\textit{et al.}~\cite{Arnold2018} used a MIMO orthogonal frequency division multiplex (OFDM) system for localization. By pre-training the neural network (NN) with LoS data, they significantly reduced the number of samples needed to achieve a good training result. 

Finally, in~\cite{Foliadis2021} Foliadis~\textit{et al.} showed methods that allow reliable wireless fingerprinting when inconsistencies with the raw phase make CSI unreliable. They proved that developing a model that pooled over subcarriers rather than antennas was more appropriate, so in our architecture we also pool over subcarriers using a kernel of size 1\,$\times$\,4.

\subsection{Carbon footprint}
\label{subsec:CO2}
In~\cite{Gigi2020}, Hsueh analyzed the carbon footprint of machine learning algorithms and concluded that convolutional layers are power hungry because they operate in three dimensions, as opposed to fully connected layers which operate in two dimensions. The model with the fewest parameters (weights) showed the best trade-off between performance and carbon footprint. In~\cite{Verhelst2017}, Verhelst~\textit{et al.} analyzed the complexity of CNNs and discussed hardware optimization techniques, mainly targeting the Internet of Things (IoT) and embedded devices.

In~\cite{jurj2020environmentally}, Jurj~\textit{et al.} proposed four different metrics that account for different aspects of the trade-off between model performance and energy consumption, while in~\cite{Garcia2019} Garcia~\textit{et al.} surveyed the energy consumption of various models. They proposed a taxonomy of power estimation models at the software and hardware levels and discussed existing approaches for estimating energy consumption. They stated that using the number of weights is not accurate enough and therefore calculating the number of floating-point operations (FLOPs) or multiply-accumulate operations (MACs) is required for accurate calculation of energy consumption. In our work, we also evaluate the carbon footprint of the compared models. 
\begin{figure*}[htbp]
    \centering
    \includegraphics[width=\linewidth]{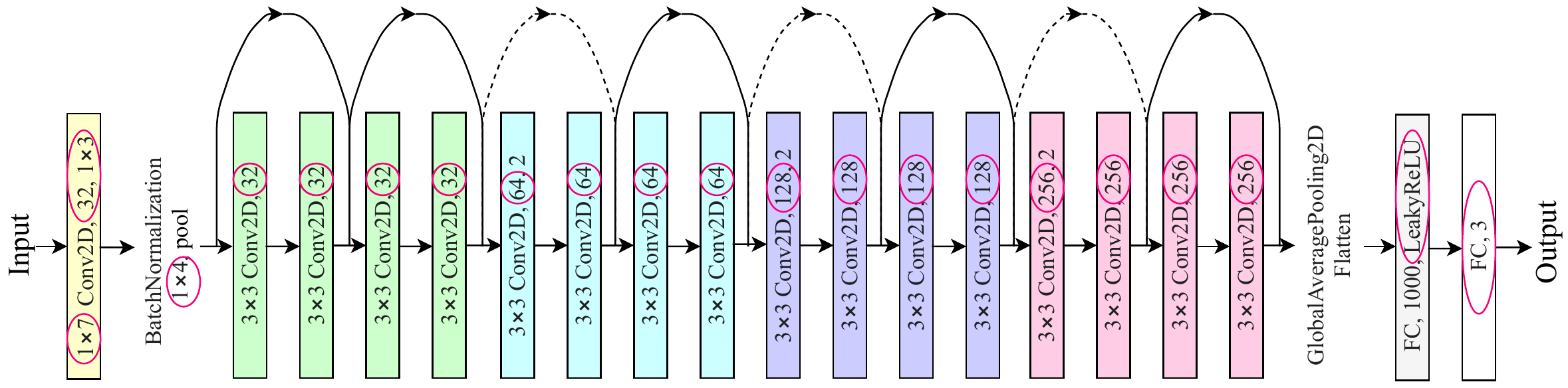}
    \caption{The proposed PirnatEco architecture adapted from ResNet18 with differences marked with red circles.}
    \label{fig:Pirnat_1G}
\end{figure*}

\section{Problem Statement}
\label{sec:setup}
Assuming a system with an antenna array of size $N \times M$ receiving transmissions from an entity E located at a position p (x, y, z), we want to develop a model capable of estimating the spatial coordinates $\tilde{p}$ ($\tilde{x}$, $\tilde{y}$, $\tilde{z}$) of E.


The model should be developed to predict the position $\tilde{p}$ of E as accurately as possible. We use two standard metrics to measure the distance between the actual position $p$ and the estimated position $\tilde{p}$, namely the mean distance error $MDE = E[||p-\tilde{p}||_2]$ and the root mean square error $RMSE = \sqrt{E[||p-\tilde{p}||^2_2]}$.

At the same time, the model should consume less energy than the state of the art models, subject to a minor performance degradation. To estimate the energy consumption for training a DL model, it is necessary to consider the number of FLOPs per type of layer used in the model architecture~\cite{Garcia2019}. 

\section{Proposed DL Network Architecture}
\label{sec:model}

To achieve high localization performance and reduce energy consumption during training and operation of a DL model, we propose a  multilayer model PirnatEco inspired by ResNet18~\cite{he2016deep}, shown in Figure~\ref{fig:Pirnat_1G}. We chose ResNet18 because it is the least complex ResNet DL model and is more adaptable to less complex types of images constructed from time series, as is the case with localization. In Figure~\ref{fig:Pirnat_1G}, each layer is visible and explained with its kernel size, type, number of nodes and in some cases stride and activation function. The red circles mark the differences with ResNet18. Unlike ResNet18, in PirnatEco the first layer is a convolutional 2D layer (Conv2D) with a kernel size of $1\times7$ and a stride of $1\times3$, followed by a batch normalization and  pooling layer with a pool size of $1\times4$. These kernels and pools are designed to move across the subcarriers of a single antenna, which is different from the square kernels and pools in ResNet18. 

Next, we use adapted ResNet blocks with reduced number of weights, where the number of nodes doubles every four layers from 32 to 256, unlike ResNet18 which starts with 64. The kernel size in the blocks is $3\times3$, similar to ResNet18. Finally, PirnatEco uses LeakyReLU activation with a parameter alpha set to 10\textsuperscript{-3} at the fully connected (FC) layer with 1000 nodes, unlike ResNet18 which uses ReLU. 

\subsection{Methodology for calculating model complexity}
Starting from the existing methods for calculating model complexity ~\cite{Verhelst2017}\footnote{https://cs231n.github.io/convolutional-networks/\#conv}, we use the following equations to calculate the FLOPs for the layers and then the total FLOPs used by PirnatEco. 

\subsubsection{Fully connected layer}
A fully connected (F$_\text{fc}$) layer performs MAC operations. Their number depends on the input size $I_\text{s}$ and the output size $O_\text{s}$. A MAC consists of 2 FLOPs. For layers that use rectifying linear units (RelU), the output size has to be added to the results of the product, as shown in Eq.~\ref{eq:FLOPSFC}.
\begin{equation}
\label{eq:FLOPSFC}
F_\text{pe} = 2\,I_\text{s}O_\text{s} + O_\text{s}
\end{equation}

\subsubsection{Convolutional layer}
A convolutional layer consists of a set of filters of size $K_\text{r} \times K_\text{c}$ used to scan an input tensor of size $I_\text{r} \times I_\text{c} \times C$ with a stride $S$. More precisely, the number of all FLOPs per filter $F_\text{pf}$ is given by Eq.~\ref{eqn:FlopsPerFilter}.
\begin{equation}
\label{eqn:FlopsPerFilter}
F_\text{pf} = (\frac{I_\text{r} - K_\text{r} + 2P_\text{r}}{S_\text{r}} + 1)(\frac{I_\text{c} - K_\text{c} + 2P_\text{c}}{S_\text{c}} + 1) (2CK_\text{r}K_\text{c} + 1)
\end{equation}

The first term of the equation gives the height of the output tensor, where $I_\text{r}$ is the size of the input rows, $K_\text{r}$ is the height of the filter, $P_\text{r}$ is the padding and $S_\text{r}$ is the size of the stride. The second term represents the same calculation for the width of the output tensor, where the indices in  $I_\text{c}$, $K_\text{c}$, $P_\text{c}$ and $S_\text{c}$ correspond to the input columns. The last term provides the number of computations per filter for each of the input channels $C$ that represent the depth of the input tensor and the bias. 

The number of FLOPs used throughout the convolutional layer is equal to the number of filters times the flops per filter given in Eq.~\ref{eqn:FlopsPerFilter}, i.e. $F_\text{c}=(F_\text{pf} + N_\text{ipf})N_\text{f}$. However, in the case where ReLU are used, one additional comparison and multiplication are required to calculate the number of FLOPs used in one epoch $F_\text{pe}$. We therefore added the number of FLOPs used for each filter and the number of instances for each filter and then multiplied by the number of all filters $N_\text{f}$:  
\begin{equation}
\label{eqn:FlopsLayerRelu}
F_\text{c} = (F_\text{pf} + (2CK_\text{r}K_\text{c} + 1))N_\text{f} .
\end{equation}

\subsubsection{Pooling layer} The pooling layer is responsible for downsampling the height and width of the input tensor. No padding is performed when pooling, and there is only one filter in it, therefore the number of FLOPs per pooling layer $F_\text{p}$ is given by:
\begin{equation}
\label{eqn:FlopsPerPool}
F_\text{p} = (\frac{I_\text{r} - K_\text{r}}{S_\text{r}} +1)(\frac{I_\text{c} - K_\text{c}}{S_\text{c}} + 1) (2CK_\text{r}K_\text{c} + 1)
\end{equation}

\subsubsection{Final model}
The process of training the model involves sequential forward and backward propagation through the different layers of the architecture. During forward propagation, the network computes the loss based on the initialized weights. During backward propagation, it updates the weights and biases based on the gradients it computed against the loss. Training is carried out in epochs, where an epoch involves going forward and then backward through all available training samples. Prediction, on the other hand, requires only one forward pass through the network.

The number of operations in a DL architecture depends on the number and types of layers $L$ and can be computed as:
\begin{equation}
\label{eqn:ModelFlops}
M_{FLOPs} = \sum_{l=1}^{L} F_\text{l} ,
\end{equation}

where $F_\text{l}$ refers to the $l\textsuperscript{th}$ layer of the architecture and corresponds to a fully connected $F_{pe}$, convolutional $F_{c}$ or pooling $F_{p}$ layer. The energy consumed during the forward propagation $E(fp)$ of the training process corresponds to the energy required for making a forward pass multiplied by the size of the training data $training_{samples}$ and the number of \textit{epochs}, as shown in Eq. \ref{eqn:Forward}:
\begin{equation}
E_{fp} = \frac{M_{FLOPs} }{GPU_{performance}}(training_{samples} \times epochs)
\label{eqn:Forward}
\end{equation}

where $GPU_{performance}$ is measured in FLOPS/Watt, and FLOPS stands for FLOPs per second.  Computing the energy for backward propagation is a more challenging step, so we approximate it as $E_{bp} = 2 \times E_{fp}$, since we know that backward propagation is generally more computationally intensive and on ResNet20 it takes about twice as long to compute as forward propagation \cite{devarakonda2017adabatch}. Therefore the energy required for training $E_{T}$ is:
\begin{equation}
E_{training} = E_{fp} + E_{bp} = 3 \times E_{fp}
\label{eqn:Training}
\end{equation}

Once we use the trained model in production, the energy required for prediction is equal to the energy required for a forward pass $E_{prediction}=M_{FLOPs}/{GPU_{performance}} \times input$, where input is the number of input samples for the prediction.

\subsection{Model training and evaluation methodology}
To develop a localization model, we used CSI and GPS measurements from the publicly available CTW\,2019 challenge\footnote{https://data.ieeemlc.org/Ds1Detail} dataset. To train and test the model, we generated four different evaluation datasets with different splits between training and testing data areas in a $9:1$ ratio, labelling the obtained evaluation sets as Random, Narrow, Wide and Within as in \cite{cerar2021improving}.
Thus, we trained and tested the model with 15723 and 1748 samples in batches of 32 samples, respectively. In each epoch, we went through $17471\times32$ samples. Weights were updated using stochastic gradient descent (SGD) with a learning rate of 0.01 and momentum of 0.9. We also ran tests with other learning rates (i.e. 0.04, 0.02, 0.005) and moments and selected the best values.

When calculating the computational cost of model training $E_{training}$  (Eq.~\ref{eqn:Training}) and operation $E_{prediction}$, we considered the number of FLOPS per watt of power of the NvidiaT4 graphics cards, since they are used by Google Colab\footnote{https://colab.research.google.com/}, on which we conducted our research. Furthermore, we calculate the carbon footprint assuming that electricity is produced with a footprint of 250\,g of CO$_\text{2}$ equivalent per kilowatt hour, as determined for the west coast of the USA from electricitymap.org.

\begin{figure}[htbp]
    \centering
    \includegraphics[width=\linewidth]{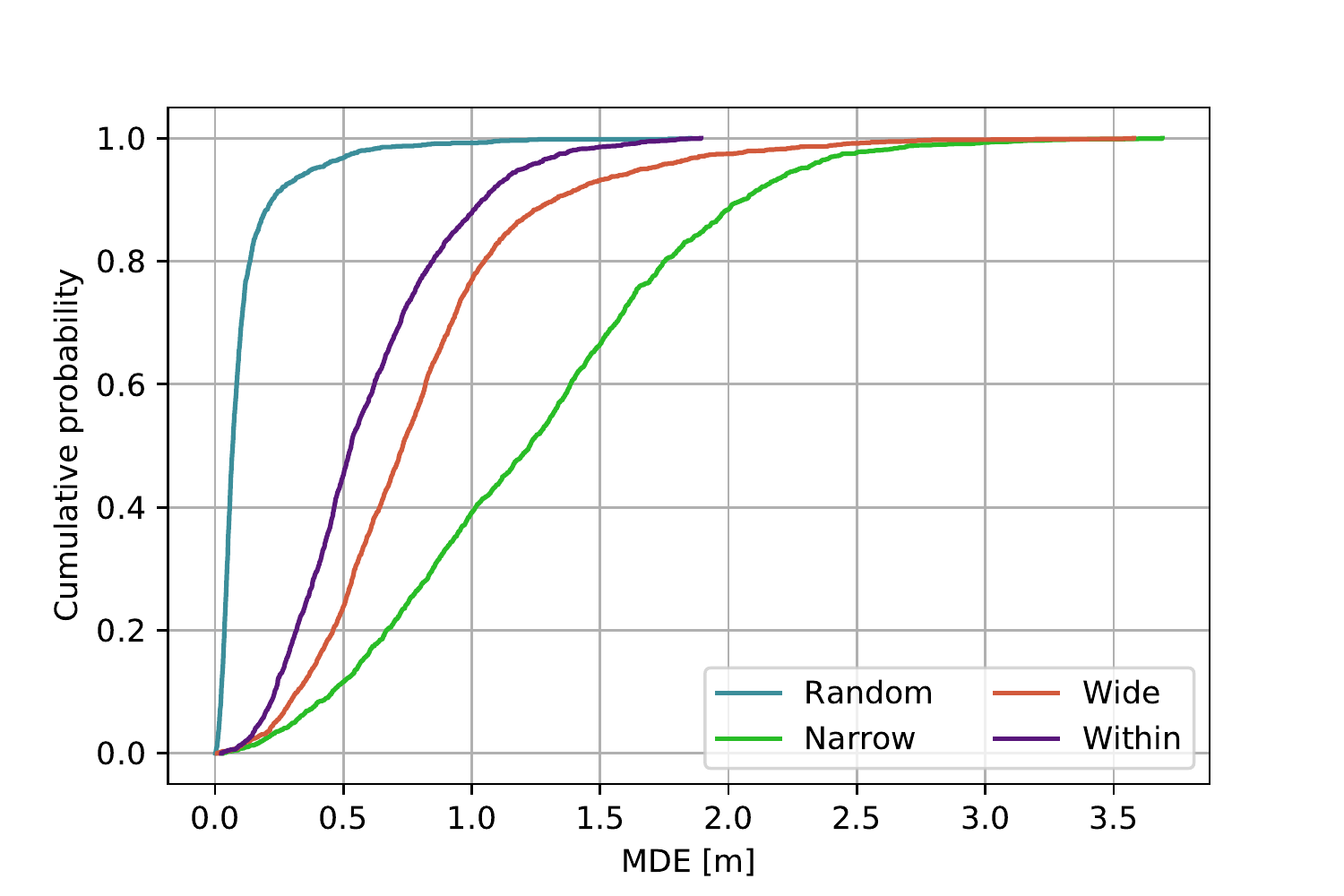}
    \caption{CDF of PirnatEco developed according to the four train/test data set splits: Random, Wide, Narrow, Within.}
    \label{fig:CDF}
    \includegraphics[width=\linewidth]{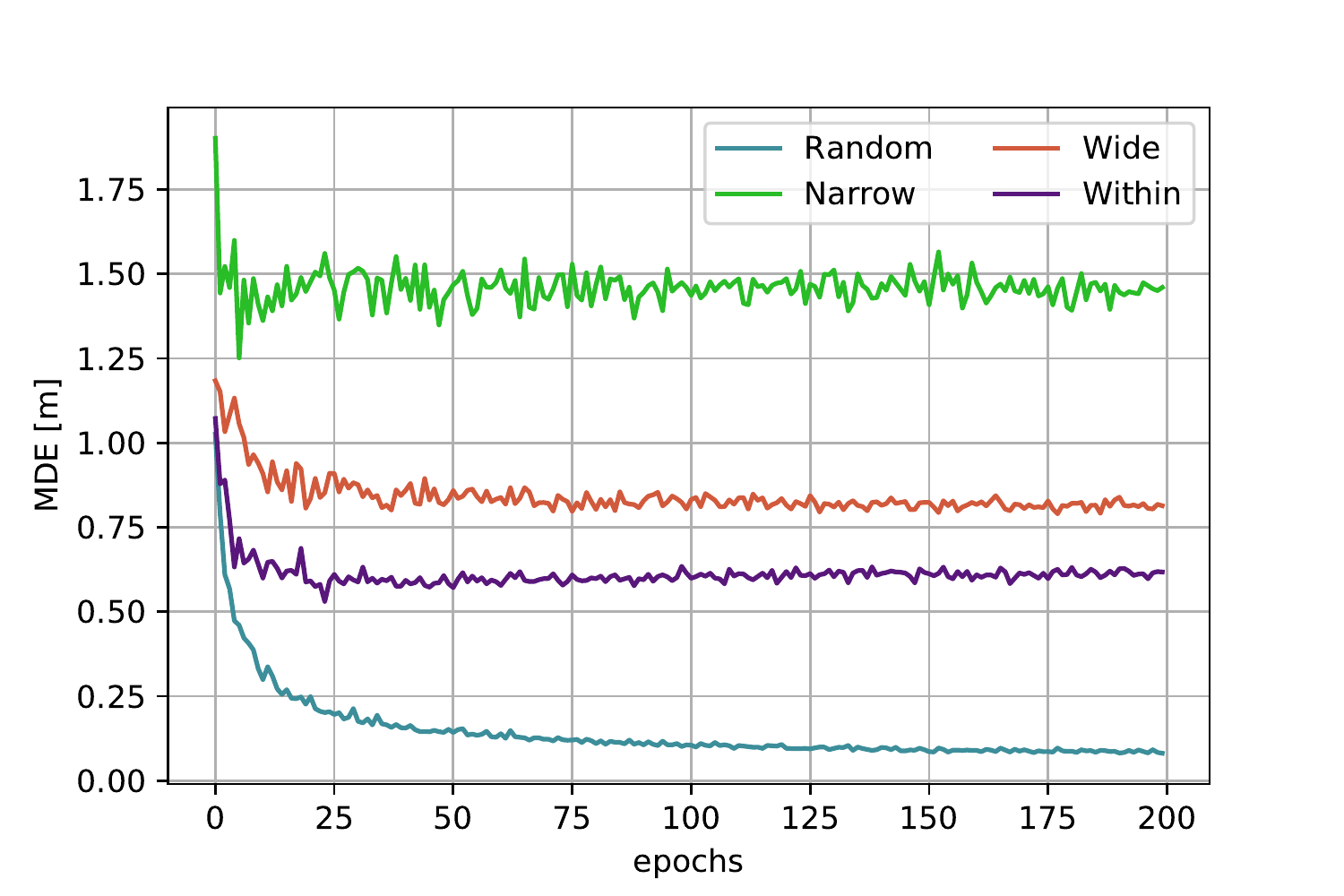}
    \caption{Performance vs. epochs for the four train/test data set splits: Random, Wide, Narrow, Within.}
    \label{fig:RNWW}
\end{figure}

\begin{figure*}[tbh]
    \centering
    \subfloat[Random: histogram\label{fig:Random_h}]{\includegraphics[width=0.25\linewidth]{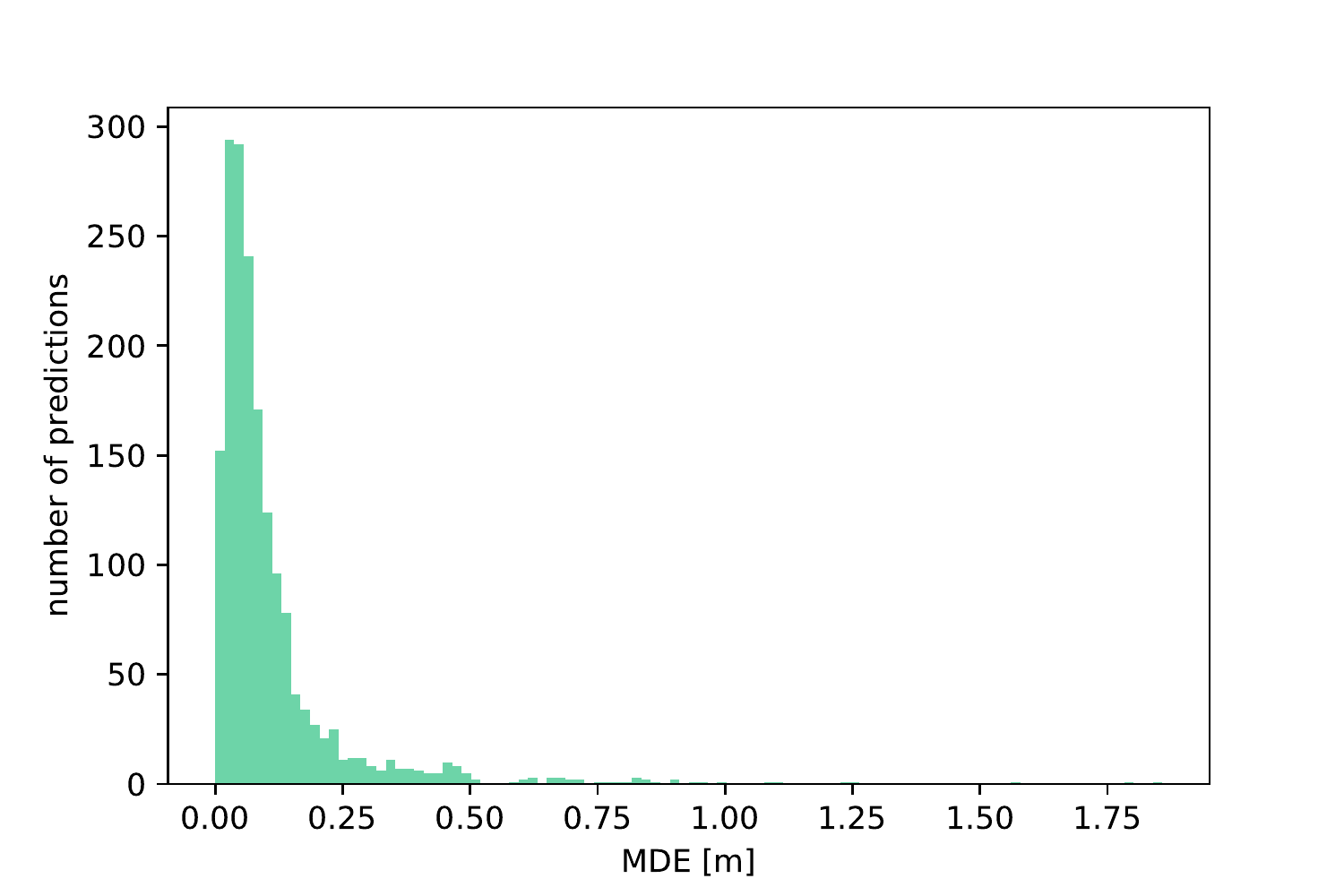}}
    \subfloat[Narrow: histogram\label{fig:Narrow_h}]{\includegraphics[width=0.25\linewidth]{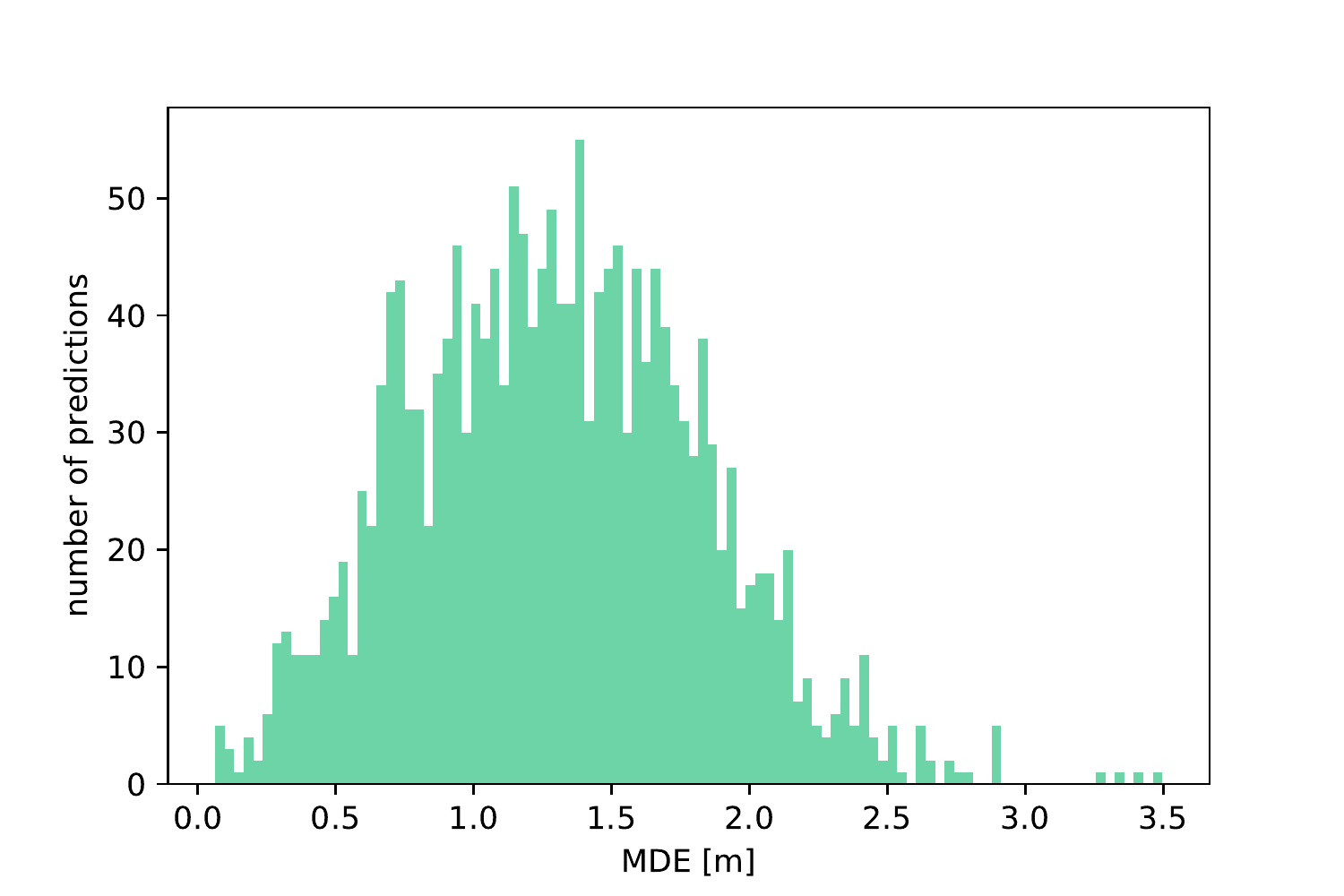}}
    \subfloat[Wide: histogram\label{fig:Wide_h}]{\includegraphics[width=0.25\linewidth]{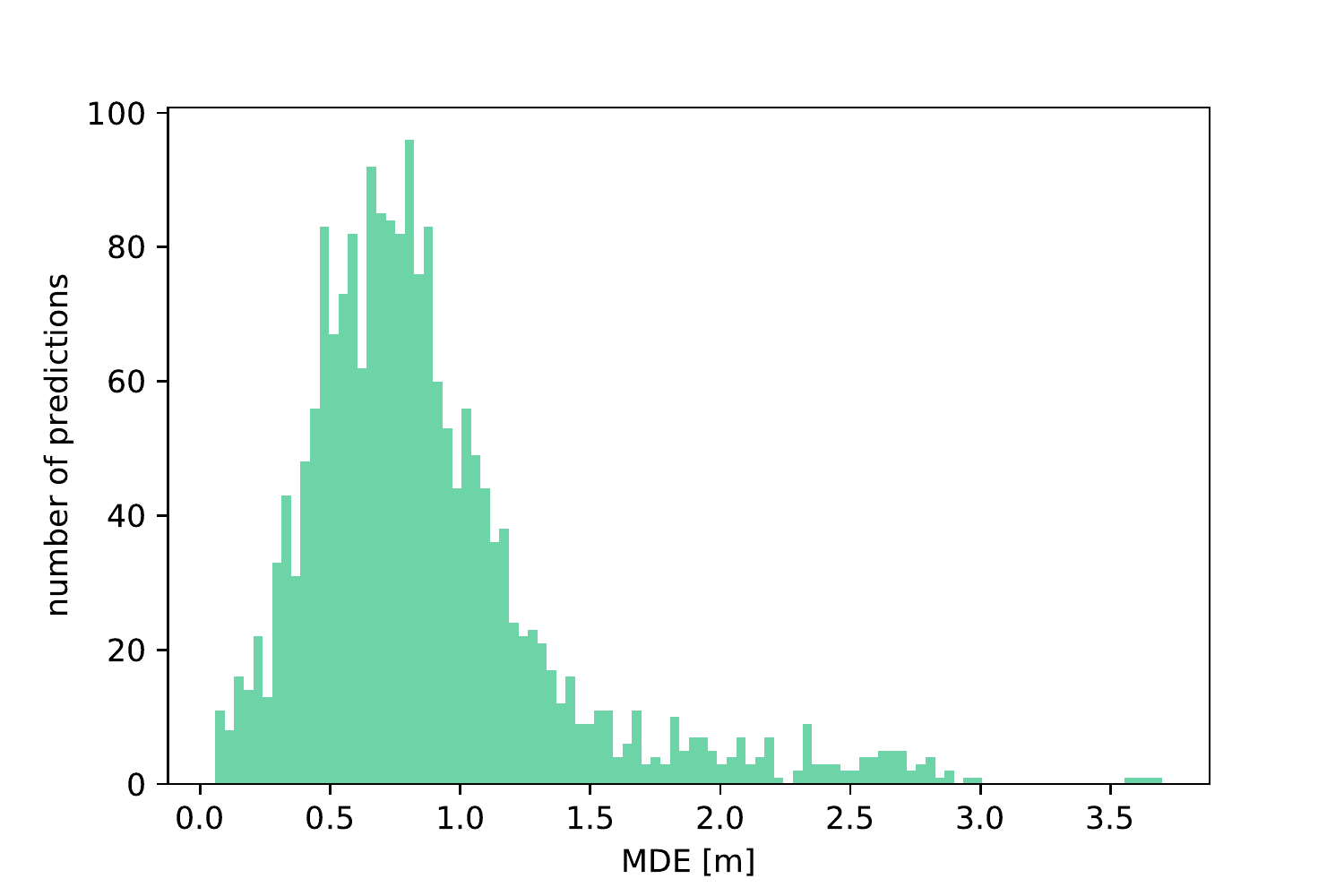}}
    \subfloat[Within: histogram\label{fig:Within_h}]{\includegraphics[width=0.25\linewidth]{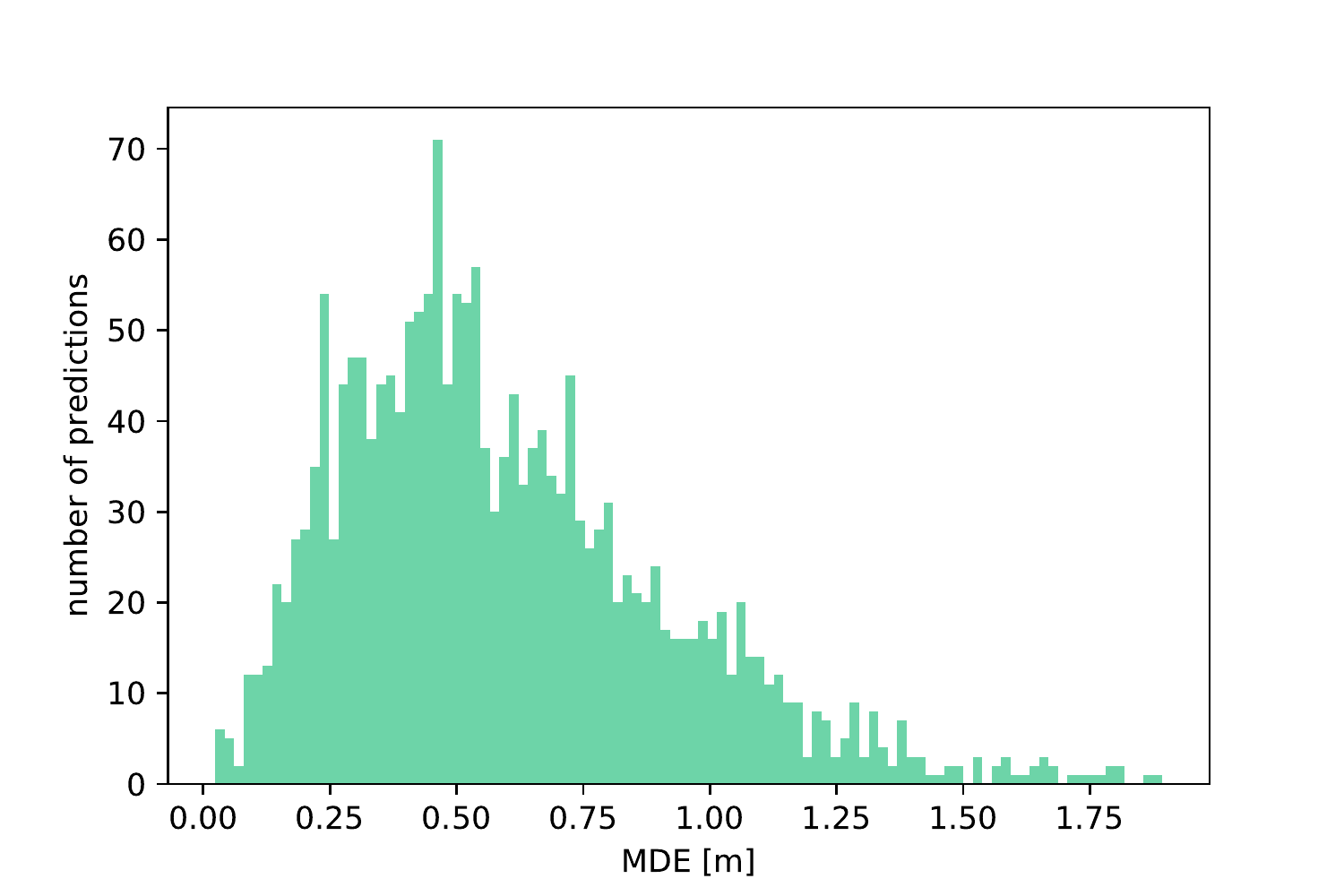}}
    \caption{Distribution of the prediction with PirnatEco.}
    \label{fig:histograms}
\end{figure*}

\begin{table*}[htb]
\ra{1.3}
\caption{Performance evaluation on CTW\,2019 dataset}
\label{tab:evaluation}
\centering
\begin{tabular}{@{}l|r|rr|rr|rr|rr@{}}
    \toprule
    \multirow{2}{*}[-0.5em]{Approach}
    & \multirow{2}{*}[0.5em]{Weights}
    & \multicolumn{2}{c|}{Random}
    & \multicolumn{2}{c|}{Narrow}
    & \multicolumn{2}{c|}{Wide}
    & \multicolumn{2}{c}{Within}
    \\
    
    \cmidrule(lr){3-4}
    \cmidrule(lr){5-6}
    \cmidrule(lr){7-8}
    \cmidrule(l){9-10}
    
    {} & $[10^6]$
    & RMSE & MDE 
    & RMSE & MDE 
    & RMSE & MDE 
    & RMSE & MDE 
    \\
    \midrule
    
    
    Dummy (linear), FCNN
    & <0.1
    & 0.724 & 1.122 
    & 1.055 & 1.809 
    & 0.878 & 1.428 
    & 0.441 & 0.721 
    \\
    
    Arnold~\etal~\cite{Arnold2018}, FCNN
    & 32.3
    & 0.570 & 0.853 
    & 1.001 & 1.594 
    & 0.733 & 1.145 
    & 0.381 & 0.584 
    \\
    
    Arnold~\etal~\cite{Arnold2018}, CNN
    & 7.6
    & 0.315 & 0.445 
    & 0.857 & 1.330 
    & 0.605 & 0.923 
    & 0.454 & 0.702 
    \\
    
    De Bast~\etal~\cite{DeBast2020}, CNN
    & 0.4
    & 0.722 & 1.120 
    & 1.110 & 1.907 
    & 0.828 & 1.331 
    & 0.377 & 0.611 
    \\
    
    Chin~\etal~\cite{chin2020intelligent} FCNN
    & 123.6
    & 0.563 & 0.838 
    & 1.007 & 1.611 
    & 0.726 & 1.133 
    & 0.365 & 0.574 
    \\
    
    Chin~\etal~\cite{chin2020intelligent} CNN
    & 13.7
    & 0.100 & 0.093  
    & 0.854 & 1.326  
    & 0.530 & 0.808  
    & 0.381 & 0.620  
    \\
    
    
     Cerar~\etal~\cite{cerar2021improving} CNN4 
    & 5.3
    & 0.122 & 0.149  
    & 0.819 & 1.286  
    & 0.514 & 0.787  
    & 0.365 & 0.552  
    \\
    
    Cerar~\etal~\cite{cerar2021improving} CNN4R 
    & 10.8
    & 0.113 & 0.127  
    & 0.776 & 1.227  
    & 0.539 & 0.835  
    & 0.351 & 0.521  
    \\

    Cerar~\etal~\cite{cerar2021improving} CNN4S 
    & 16.3
    & 0.108 & 0.120  
    & 0.821 & 1.285  
    & 0.528 & 0.804  
    & 0.351 & 0.524  
    \\
    \midrule
    PirnatEco
    & 3.1
    & 0.109 & 0.112  
    & 0.801 & 1.260  
    & 0.523 & 0.793  
    & 0.398 & 0.596  
    \\
    
\bottomrule
\end{tabular}
\end{table*}
\section{Performance Evaluation}
\label{sec:evaluation}
To evaluate the proposed PirnatEco model, we first evaluate the performance of the model and then quantify the energy consumption for its training and prediction.

\subsection{Performance of the PirnatEco model}
Figure~\ref{fig:CDF} shows the performance of the model using a cumulative distribution function (CDF) of the MDE of the estimated position $\tilde{p}$~\textit{x}. It can be seen that a very large majority of the locations predicted for the Random category have an accuracy of 0-0.2\,m, providing the best performance. This is followed by the Within and Wide categories, where the accuracy for most locations is in the range of 0.1-1\,m and 0.2-1.5\,m, respectively. The worst performance is obtained for the Narrow category, where the prediction accuracy for 90\,\% of locations only reaches 0.5-2\,m. 

To select the best model for each of the four evaluation sets, we evaluated accuracy as a function of epochs, as shown in Figure~\ref{fig:RNWW}. It can be seen that the performance improvement slows down after 85 epochs for the Random category and after 15 epochs for the Wide category. Accuracy for the Narrow category shows the worst results, with no obvious relation between accuracy and epochs, while for the Within category the best performance is obtained after 20 epochs and slightly deteriorates after 50 epochs. Considering these results, we used 85 epochs in the Random category, 15 epochs in the Narrow and Wide categories, and 20 epochs in the Within category. For comparison Chin~\etal~\cite{chin2020intelligent} model needed 67, 30, 23 and 31 epochs, while Cerar~\etal~\cite{cerar2021improving} needed 181, 32, 34 and 68 epochs, respectively.

Further insight into the quality of the proposed model is provided by the histograms in Figure~\ref{fig:histograms}, depicting the distribution of predictions as a function of MDE for different dataset splits. In the case of Random, the spread of MSE values is very narrow around very small values and shows high accuracy. In the case of Narrow, the dispersion is relatively large and forms a bell around 1.2\,m. In the case of Wide, the bell is narrower and higher around 0.8\,m with relatively few outliers above 1.5\,m, while in the case of Within the spread is relatively large but still with most values below 1\,m.

\subsection{Performance comparison with the state of the art}
Table~\ref{tab:evaluation} summarizes different proposed localization models in terms of number of weights and accuracy in the four considered evaluation categories. 
As explained in the previous subsection and also evident in the table, the most accurate localization was obtained using the Random category, where the training and test samples are much closer to each other and both distributed across the entire area of interest, thus reducing the effects of unbalanced training. The difference in the success of the neural network structures compared is quite large. The worst performing model, De Bast~\etal~\cite{DeBast2020}, is more than a meter away from PirnatEco, but also has 7.75 times fewer weights.
The best performing results are less than 2\,cm away from ours and have at least four times more weights.

However the results were not as far apart in other evaluation categories. Our structure did not perform as well in the Within category, which was second in the overall localization accuracy achieved. We believe that this can also be explained by the aforementioned logic of balanced and unbalanced training. In this category, the differences were actually the smallest, and all models achieved accuracy within the range of 20\,cm.

The worst results were obtained for the Narrow category, which had the largest difference between training and test datasets, followed by the Wide category with slightly more balanced training, but which still did not produce as accurate results as the Within or Random categories. However, our model was among the best performing also in the Narrow and Wide categories. 


\begin{table}[htb]
    \ra{1.3}
    \caption{CO$_2$ footprint used in training}
    \label{tab:footprint}
    \centering
    \begin{tabular}{l|c|c|c}
        \toprule
        \bfseries NN & \bfseries carbon footprint & \bfseries FLOPs  & \bfseries energy \\
        \midrule
        PirnatEco
        & 10.6\,g\,CO$_2$\,eq.
        & 345\,$\cdot\,10^6$
        & 152\,kJ\\
        Chin~\etal~\cite{chin2020intelligent} CNN 
        & 18.3\,g\,CO$_2$\,eq. 
        & 535\,$\cdot\,10^6$
        & 264\,kJ\\
        Cerar~\etal~\cite{cerar2021improving} CNN4R 
        & 176.9\,g\,CO$_2$\,eq. 
        & 2479\,$\cdot\,10^6$ 
        & 2547\,kJ\\
    \bottomrule
    \end{tabular}
\end{table}

\subsection{Environmental costs for training and prediction}
Finally, we also evaluated the best performing models from Table~\ref{tab:evaluation} in terms of the carbon footprint for their training. The calculated carbon footprints for the selected models are summarized in Table~\ref{tab:footprint}. The results represent an average energy consumption and carbon footprint needed for training a model for one of the four presented categories. The results show that on average PirnatEco produces only 6\,\% of the carbon footprint of CerarCNN4 and 58\,\% of  ChinCNN, while their performance is very comparable, i.e. our model achieves 99.4\,\% of the performance of CerarCNN4 and 98.7\,\% of the performance of ChinCNN.

\begin{figure}[htb]
    \centering
    \includegraphics[width=\linewidth]{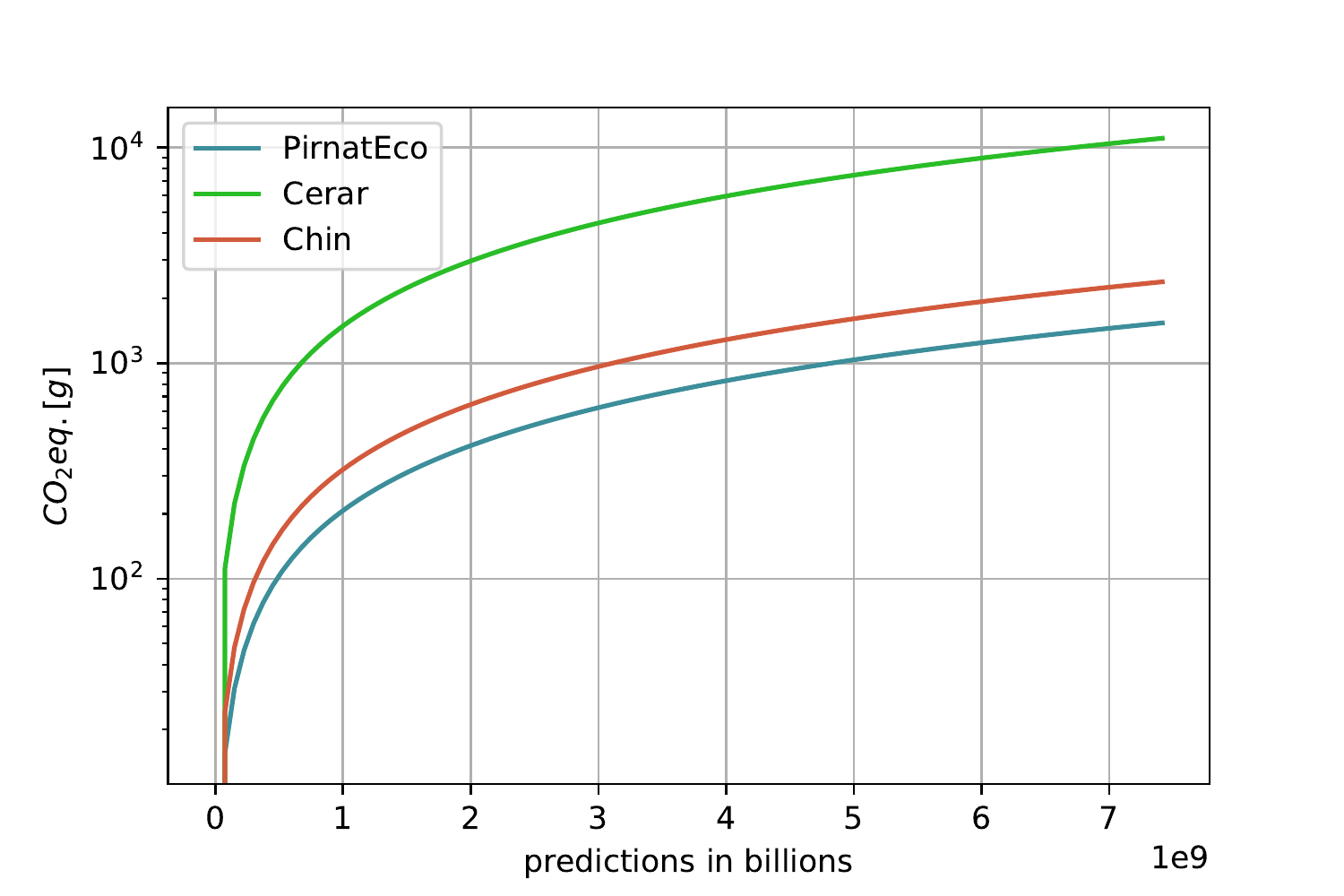}
    \caption{Carbon footprint vs predictions made in logarithmic scale}
    \label{fig:CTPL}
\end{figure}

In Figure~\ref{fig:CTPL}, we plot the calculated CO$_2$ emissions as a function of the number of location predictions. The final number in the graph shows CO$_2$ emissions produced if we made only one prediction for each mobile user in 2025 when the estimate number of mobile users is supposed to exceed 7.4 billion.

\section{Conclusions}
\label{sec:conclusion}
In this paper, we propose a new DL architecture used in the PirnatEco model for indoor positioning, paying special attention to energy efficiency during training and operation with only minor performance degradation compared to similar models. In developing the architecture, we started from the ResNet18 architecture and (i) reduced the size of the filters and (ii) adapted the pools, while being aware of the specificities of the data available for the problem. Since there is a paucity of work evaluating the energy efficiency and computational complexity of DL models, we also elaborated the methodology to benchmark the three best performing models in terms of their carbon footprint for training and prediction. We have shown that it is possible to develop DL models for wireless fingerprinting localization that optimize both accuracy and environmental cost, providing a viable alternative to models that focus only on accuracy.






\section*{Acknowledgments}
This work was funded in part by the Slovenian Research Agency under the grant P2-0016.

\ifCLASSOPTIONcaptionsoff
  \newpage
\fi

\bibliographystyle{IEEEtran}
\bibliography{bibliography}

\end{document}